\pdfoutput=1

\documentclass[11pt]{article}

\usepackage{coling}

\usepackage{times}
\usepackage{latexsym}

\usepackage[T1]{fontenc}

\usepackage[utf8]{inputenc}

\usepackage{microtype}

\usepackage{inconsolata}

\usepackage{graphicx}

\usepackage{multirow}
\usepackage{tabularx}
\usepackage{adjustbox}
\usepackage{subcaption}
\usepackage{wrapfig}
\usepackage{placeins}
\usepackage{amsmath}
\usepackage{float}
\usepackage{placeins}

\usepackage{microtype}
\usepackage{hyperref}
\usepackage{url}
\usepackage{booktabs}

\usepackage{xcolor}
\usepackage{colortbl}
\usepackage{dcolumn}
\usepackage{pdflscape}
\usepackage{rotating}

\definecolor{darkblue}{rgb}{0, 0, 0.5}
\hypersetup{colorlinks=true, citecolor=darkblue, linkcolor=darkblue, urlcolor=darkblue}

\graphicspath{{figures/}}

\title{FTFT: Efficient and Robust Fine-Tuning by Transferring Training Dynamics}

\author{Yupei Du, Albert Gatt, and Dong Nguyen \\
  Utrecht University \\
  Utrecht, the Netherlands \\
  \texttt{\{y.du, a.gatt, d.p.nguyen\}@uu.nl}}

\begin{document}
\maketitle
\maketitle

\begin{abstract}
Despite the massive success of fine-tuning Pre-trained Language Models (PLMs), 
they remain susceptible to out-of-distribution input. 
Dataset cartography is a simple yet effective dual-model approach 
that improves the robustness of fine-tuned PLMs. It involves 
fine-tuning a model on the original training set (i.e. reference model), 
selecting a subset of important training instances  
based on the training dynamics, 
and fine-tuning again only on these selected examples (i.e. main model).
However, this approach requires fine-tuning the same model twice, 
which is computationally expensive for large PLMs. 
In this paper, we show that
1) training dynamics are highly transferable across model sizes 
and pre-training methods, and that
2) fine-tuning main models using these selected training instances 
achieves higher training efficiency than empirical risk minimization (ERM).
Building on these observations,
we propose a novel fine-tuning approach:
Fine-Tuning by transFerring Training dynamics (FTFT). 
Compared with dataset cartography, 
FTFT uses more efficient reference models and aggressive early stopping. 
FTFT achieves robustness improvements over ERM 
while lowering the training cost by up to $\sim 50\%$.\footnote{
    Our code is publicly available at \url{https://github.com/nlpsoc/FTFT}. 
    
}
\end{abstract}

\section{Introduction}

Despite the success of few-shot and zero-shot learning~\citep{brown2020language},
state-of-the-art performance in Natural Language Processing (NLP) is still largely achieved by 
fine-tuning large Pre-trained Language Models (PLMs)~\citep{mosbach-etal-2023-shot}.
Scaling laws \citep{kaplan2020scaling,hoffmann2022training} suggest that 
better downstream performance is achieved with larger PLMs.
However, fine-tuning large PLMs is also more expensive, 
in terms of both computational resources and carbon emission \citep{strubell-etal-2019-energy,MLSYS2022_462211f6}.

Moreover, despite impressive progress on regular benchmarks, 
many studies have shown that 
fine-tuned PLMs lack robustness against out-of-distribution (OOD) input. 
For instance, 
human annotators can easily exploit the weaknesses of fine-tuned PLMs 
to trick these models to yield incorrect predictions, 
on tasks such as Natural Language Inference (NLI) \citep{nie-etal-2020-adversarial} 
and Hate Speech Detection (HSD) \citep{vidgen-etal-2021-learning}. 

The problem of robustness can be mitigated using dual-model approaches.
With such approaches, first a \textbf{reference model} is trained
to estimate the importance of each training instance, 
and then a \textbf{main model} is trained based on the outputs of the reference model 
\citep{lff,utama-etal-2020-mind,poe_pipeline_agnostic,karimi-mahabadi-etal-2020-end,cnc,jtt}. 
Among these approaches, 
dataset cartography \citep{swayamdipta-etal-2020-dataset} is especially attractive 
in view of its simplicity and the consistent improvements in model robustness. 
It consists of three steps. 
First, a \textbf{Data Map (DM)} is constructed 
from the \textbf{training dynamics} (i.e. instance prediction probabilities) 
of a fine-tuning run of the reference model on the full dataset.
This DM divides the training data into three subsets: 
ambiguous, hard-to-learn, and easy instances. 
Finally, the main model is fine-tuned 
using only either the ambiguous or hard-to-learn subset. 
An important question has to do with the choice of reference model. 
\citet{swayamdipta-etal-2020-dataset} use the same PLM for both the reference and main model. 
However, a major drawback of dataset cartography is the high computational cost,
because it requires fine-tuning the same model twice. 
In this paper, 
\emph{we jointly address robustness and efficiency issues} 
without sacrificing the simplicity of dataset cartography, 
by exploiting the \textbf{transferability of training dynamics}. 
We make three contributions.

First, we study the following question: 
Are DMs transferable across different model sizes and pretraining methods?
We focus on the novel setting where 
\emph{DMs are constructed based on computationally efficient reference models 
to fine-tune more capable --- and often larger --- main models}.
Our motivation is two-fold: 
1) efficient reference models ease the computational burden of constructing DMs, 
and 2) less capable reference models might be better  
at identifying ambiguous or hard training instances, 
because they are less likely to memorize training data \citep{tirumala2022memorization,carlini2023quantifying}. 
Our results show that, in most cases, 
\emph{training dynamics are highly transferable} across different model sizes 
(\S\ref{subsec:transferability_across_sizes})
and pretraining methods~(\S\ref{subsec:transferability_across_model_architectures}). 
We further show that the condition for successful transfers is 
a reasonably strong reference model, 
which we also make precise in \S\ref{subsec:how_efficient_can_we_be}. 

Second, we observe that
fine-tuning with selected instances 
achieves consistently higher training efficiency than conventional fine-tuning 
(\S\ref{sec:ftft}).

Third, building on these findings, 
we propose \textbf{Fine-Tuning by transFerring Training dynamics (FTFT, \S\ref{sec:ftft})}:
an efficient fine-tuning approach that leads to improved OOD performance.
Compared to dataset cartography, 
FTFT uses \emph{efficient reference models} and \emph{early stopping}. 
Experiments on two tasks, NLI and HSD, 
show that FTFT achieves better performance on OOD input 
than conventional Empirical Risk Minimization (ERM), 
while lowering the training cost by up to $\sim 50\%$.

\section{Background}

\paragraph*{Dual-Model Approaches for Robustness}

Many studies have proposed dual-model approaches to improve model robustness 
that do not require knowledge of identifiable subsets of the data samples, 
such as NLI pairs in which hypotheses contain a negation~\citep{gururangan-etal-2018-annotation}, 
or HSD samples targetting a specific group~\citep{dixon2018measuring,park-etal-2018-reducing}. 
\citet{lff} first train a reference model using generalized cross-entropy loss, 
and then train a main model while assigning higher weights to instances 
that were hard for the reference model. 
\citet{poe_pipeline_agnostic} use a Product-of-Expert (PoE) approach, 
by first training a reference model with limited capacity to capture dataset biases, 
and then training the main model to avoid these biases using PoE loss. 
\citet{jtt} propose the Just-Train-Twice approach (JTT), 
which involves first training a weak reference model 
using heavy regularization and vanilla SGD, 
and then up-weighing the training instances that the reference model predicts incorrectly %
when training the main model.
Dataset cartography \citep{swayamdipta-etal-2020-dataset} is based on a similar idea,
but it uses training dynamics instead of correctness to categorize training instances. 
We discuss this method below.

\paragraph*{Dataset Cartography} %

is a dual-model approach to improve model robustness. 
First, a reference model is trained on the full training dataset. 
Then, a Data Map (DM) is built based on the training dynamics, 
by tracking prediction probabilities of the true class ($p_{\mathrm{true}}$) 
of each training instance across epochs.
DM categorizes training instances into three subsets: 
{\em ambiguous} (i.e. the variance of $p_{\mathrm{true}}$ is in the top $q\%$ of all training instances); 
{\em hard-to-learn} (i.e. the mean of $p_{\mathrm{true}}$ is in the bottom $q\%$ of all training instances); 
and {\em easy} (i.e. neither ambiguous nor hard-to-learn). 
The threshold $q\%$ is fixed and typically set to $33\%$. 
Note that a training instance can be categorized as 
both hard-to-learn and ambiguous (a low mean but high variance for $p_{\mathrm{true}}$). 
Finally, the main model is fine-tuned only on the 
ambiguous or hard-to-learn subset. 
\citet{swayamdipta-etal-2020-dataset} show that, 
with a slight loss of In-Distribution (ID) performance, 
dataset cartography improves Out-Of-Distribution (OOD) performance of models.  
In this paper, we train main models with ambiguous data, 
since \citet{swayamdipta-etal-2020-dataset} reported better performance using this data than hard-to-learn data.

\citet{swayamdipta-etal-2020-dataset} use the same PLM as both the reference and the main model. 
In contrast, \citet{sar-shalom-schwartz-2023-curating} show that 
a DM constructed by $\text{ELECTRA}_{\text{Large}}$ \citep{clark2020electra} can be used 
to improve the robustness of $\text{DeBERTaV3}_{\text{Large}}$ \citep{he2023debertav}. 
However,
instead of using only the ambiguous subset, 
they added $k$ copies of this subset to the original training set to train the main model.
Moreover, they did not investigate either DM transfer across model sizes and pretraining methods, 
or how such transferability can be exploited to improve efficiency. 

\paragraph{Model-Based Data Selection/Reweighing}
Our work is also connected to studies that have
selected or reweighed data using reference models to improve in-distribution (ID) performance. 
\citet{NIPS2017_2f37d101} use $p_{\text{true}}$ variance and proximity to the classification threshold 
from a reference model to reweigh training instances; 
\citet{toneva2018an} calculate the frequency of forgetting events (i.e. from a correct to incorrect prediction), 
and remove the least forgettable instances; 
\citet{paul2021deep} and \citet{baldock2021deep} instead use 
error vector norm and effective prediction depth to estimate the contribution of a training instance. 

Previous studies have also explored the use of a smaller reference model to improve efficiency. 
\citet{coleman2020selection} use a small model for active learning and core-set selection.
\citet{xie2023doremi} reweigh domains for language model pretraining, by training a small reference model to estimate the difficulty of each domain.

\section{Experimental Setup}

We perform our experiments on two tasks,
Natural Language Inference (NLI) and Hate Speech Detection (HSD). 
Following \citet{swayamdipta-etal-2020-dataset}, 
we select 33\% the most ambiguous datapoints. 

\paragraph{Data}
To study model robustness, 
we include challenging OOD test sets for each task, 
in addition to the training set and an ID validation set.
For NLI, we use the MultiNLI dataset \citep{williams-etal-2018-broad} 
as the train and ID validation set, 
because of its diverse composition, covering 10 genres.
We use AdversarialNLI~\citep{nie-etal-2020-adversarial} as the OOD test set, 
which consists of three rounds of adversarial data collection. AdversarialNLI 
 is known to be challenging and it targets weaknesses of models trained on MultiNLI.
 
For HSD, we use CAD~\citep{vidgen-etal-2021-introducing} as the training and ID validation set. 
CAD consists of Reddit posts covering diverse topics and writing styles, 
annotated using a fine-grained taxonomy.
Following \citet{ramponi-tonelli-2022-features}, 
we frame it as a binary classification task, 
 marking identity-related abuse as hateful and other categories as non-hateful. 
As OOD test sets, we use DynaHate~\citep{vidgen-etal-2021-learning}, 
since it aligns with CAD's definition of hate speech.
DynaHate contains three rounds of adversarial data collection and perturbations.

\paragraph{Models}

We mainly use DeBERTaV3~\citep{he2023debertav} and ELECTRA~\citep{clark2020electra} in our experiments, 
due to their strong performance and availability in multiple model sizes.
To study the transferability across different pretraining methods, 
we also use TinyBERT~\citep{turc2020wellread}, 
BERT~\citep{devlin-etal-2019-bert} and RoBERTa~\citep{liu2019roberta} as reference models.\footnote{ 
Costs for fine-tuning different PLMs are in Appendix \ref{app:experimental_setup}.
We report FLOPs rather than GPU hours 
because we noticed occasional low GPU utilization especially when fine-tuning smaller PLMs.}
Besides ERM, we include a baseline that uses a random DM (i.e., random $q\%$ of the training data). 
We perform a search of optimal training steps on ERM, 
and use the same hyper-parameters for both ERM and data cartography approaches. 
Full details of the models and training setups are in Appendix \ref{app:experimental_setup}.

\begin{table*}[htbp]
    \small
    \centering
    \begin{adjustbox}{max width=0.95\linewidth}
    \begin{tabular}{lll|c|cccc}
        \toprule
        Method & Main Model & Ref. Model & Cost & MultiNLI & \multicolumn{3}{c}{AdversarialNLI (Test)} \\
        &  &  &  & - & R1 & R2 & R3 \\
        \midrule
        \multicolumn{8}{c}{Baselines: ERM, ERM with Early Stopping, and Random DM} \\ 
        \midrule
         ERM & $\text{DeBERTaV3}_{\text{Small}}$ & - & 14.47 & ${87.76}_{0.09}$ & ${33.25}_{1.67}$ & ${30.07}_{0.71}$ & ${31.89}_{0.46}$ \\
         ERM & $\text{DeBERTaV3}_{\text{Base}}$ & - & 28.29 & ${90.03}_{0.14}$ & ${43.73}_{0.66}$ & ${33.95}_{0.53}$ & ${33.79}_{1.20}$ \\
         ERM & $\text{DeBERTaV3}_{\text{Large}}$ & - & 100.00 & ${91.15}_{0.08}$ & ${59.90}_{1.95}$ & ${45.10}_{1.39}$ & ${42.08}_{0.95}$ \\
         ERM(ES) & $\text{DeBERTaV3}_{\text{Large}}$ & - & 66.67 & ${91.30}_{0.29}$ & ${59.20}_{1.28}$ & ${44.35}_{1.66}$ & ${40.29}_{0.98}$ \\
         DM & $\text{DeBERTaV3}_{\text{Large}}$ & Random & 100.00 & ${90.66}_{0.23}$ & ${55.05}_{0.78}$ & ${43.83}_{0.72}$ & ${39.48}_{0.43}$ \\
        \midrule
        \multicolumn{8}{c}{Training Dynamics Transferability: Across Different Model Sizes} \\ 
        \midrule
         DM & $\text{DeBERTaV3}_{\text{Large}}$ & $\text{DeBERTaV3}_{\text{Large}}$ & 200.00 & ${90.92}_{0.16}$ & ${59.02}_{1.97}$ & ${46.08}_{2.37}$ & ${41.85}_{0.30}$ \\
         DM & $\text{DeBERTaV3}_{\text{Large}}$ & $\text{DeBERTaV3}_{\text{Small}}$ & 114.47 & ${90.77}_{0.17}$ & ${60.10}_{1.58}$ & ${46.23}_{0.44}$ & ${41.46}_{1.01}$ \\
         DM & $\text{DeBERTaV3}_{\text{Large}}$ & $\text{DeBERTaV3}_{\text{Base}}$ & 128.29 & ${90.64}_{0.24}$ & ${59.67}_{1.19}$ & ${45.88}_{1.09}$ & ${43.00}_{1.56}$ \\
        \midrule
        \multicolumn{8}{c}{Training Dynamics Transferability: Across Different Pretraining Methods} \\ 
        \midrule
         DM & $\text{DeBERTaV3}_{\text{Large}}$ & $\text{ELECTRA}_{\text{Small}}$ & 104.61 & ${90.84}_{0.03}$ & ${50.68}_{1.67}$ & ${39.92}_{0.59}$ & ${37.10}_{0.94}$ \\
         DM & $\text{DeBERTaV3}_{\text{Large}}$ & $\text{ELECTRA}_{\text{Base}}$ & 136.18 & ${90.28}_{0.21}$ & ${60.75}_{1.06}$ & ${47.30}_{0.74}$ & ${42.92}_{0.94}$ \\
         DM & $\text{DeBERTaV3}_{\text{Large}}$ & $\text{RoBERTa}_{\text{Large}}$ & 216.78 & ${90.26}_{0.06}$ & ${61.02}_{1.11}$ & ${46.85}_{0.58}$ & ${42.65}_{0.89}$ \\
         DM & $\text{DeBERTaV3}_{\text{Large}}$ & $\text{BERT}_{\text{Large}}$ & 213.49 & ${89.37}_{0.15}$ & ${62.00}_{0.92}$ & ${48.60}_{0.65}$ & ${44.73}_{0.48}$ \\
        \midrule
        \multicolumn{8}{c}{FTFT: Efficient Reference Models + Aggressive Early Stopping} \\ 
        \midrule
         FTFT & $\text{DeBERTaV3}_{\text{Large}}$ & $\text{DeBERTaV3}_{\text{Small}}$ & 51.97 & ${90.74}_{0.12}$ & ${59.38}_{1.56}$ & ${45.80}_{2.55}$ & ${42.38}_{2.31}$ \\
         FTFT & $\text{DeBERTaV3}_{\text{Large}}$ & $\text{DeBERTaV3}_{\text{Base}}$ & 74.12 & ${90.54}_{0.29}$ & ${59.80}_{1.88}$ & ${45.85}_{1.48}$ & ${42.39}_{1.25}$ \\
         FTFT & $\text{DeBERTaV3}_{\text{Large}}$ & $\text{ELECTRA}_{\text{Base}}$ & 79.93 & ${90.03}_{0.57}$ & ${60.75}_{1.33}$ & ${47.35}_{2.28}$ & ${44.17}_{0.82}$ \\
        \bottomrule
    \end{tabular}
    \end{adjustbox}
    \caption{
        Our results with $\text{DeBERTaV3}$ as the main model on NLI (measured by accuracy scores), 
        which consist of four parts:
        (1) Baselines: 
            $\text{DeBERTaV3}$ of different sizes trained using ERM, 
            $\text{DeBERTaV3}_{\text{Large}}$ trained using ERM with early stopping (ERM(ES)), and 
            $\text{DeBERTaV3}_{\text{Large}}$ trained using random DM (random 33\% of the training data); 
        (2) Training dynamics transferability across different sizes: 
            training $\text{DeBERTaV3}_{\text{Large}}$ as the main model, 
            using DMs constructed by different sizes of $\text{DeBERTaV3}$ as reference models;
        (3) Training dynamics transferability across different pretraining methods:
            training $\text{DeBERTaV3}_{\text{Large}}$ as the main model, 
            using DMs constructed by different pretraining methods as reference models 
            ($\text{ELECTRA}_{\text{Small/Base}}$,
            BERT, and $\text{RoBERTa}$); 
        (4) FTFT (\S\ref{sec:ftft}): 
            training $\text{DeBERTaV3}_{\text{Large}}$ using our approach FTFT (\S\ref{sec:ftft}), 
            with DMs constructed by different reference models, as well as aggressive early stopping.
        R1--R3 in AdversarialNLI refer to different rounds of data collection. 
        Cost column contains the relative computational cost of each method, 
        compared to training $\text{DeBERTaV3}_{\text{Large}}$ using ERM.
        For example, the compute of $\text{DeBERTaV3}_{\text{Base}}$ with ERM is 28.29, 
        meaning its compute costs are 28.29\% of training $\text{DeBERTaV3}_{\text{Large}}$ using ERM. 
        We observe that: 
        (1) Training dynamics are transferrable across different sizes and pretraining methods,
        as constructing DMs using different reference models results in comparable performance; 
        (2) FTFT achieves robustness improvements over ERM, 
        while lowering the training cost by up to $\sim 50\%$.
    }
    \label{tab:nli_overall}
\end{table*}

\section{Transferability of Training Dynamics}\label{sec:transferability_of_training_dynamics}

\begin{figure*}
    \centering
    \begin{subfigure}{.33\textwidth}
        \centering
        \includegraphics[width=\linewidth]{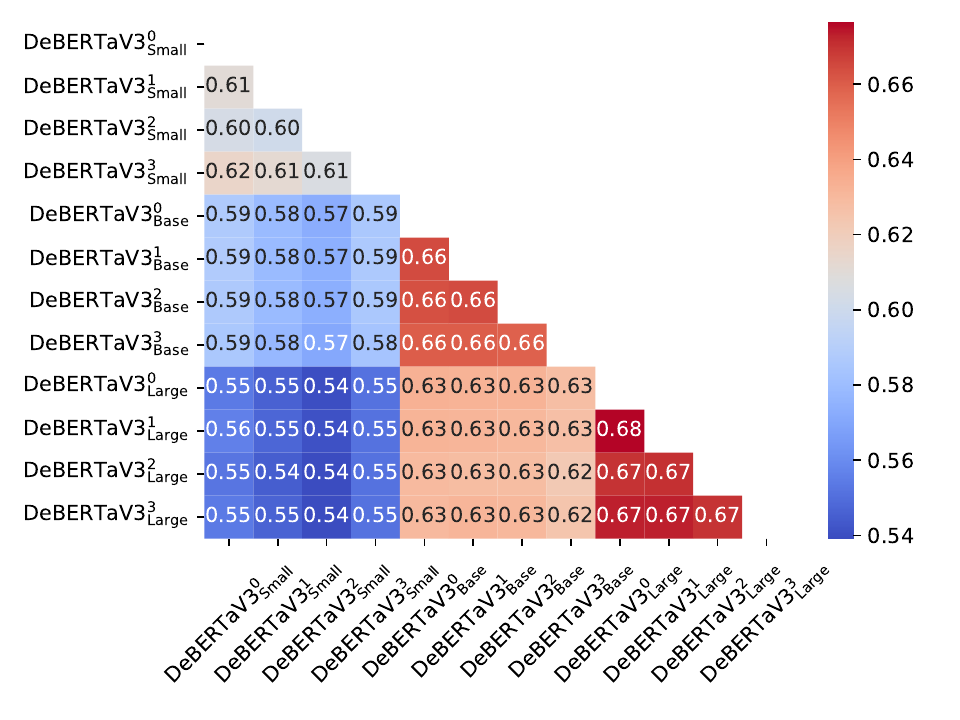}
        \caption{Consistency across different sizes}
        \label{fig:transferability_sizes_instance_level}
    \end{subfigure}
    \hfill
    \begin{subfigure}{0.31\textwidth}
        \centering
        \includegraphics[width=\linewidth]{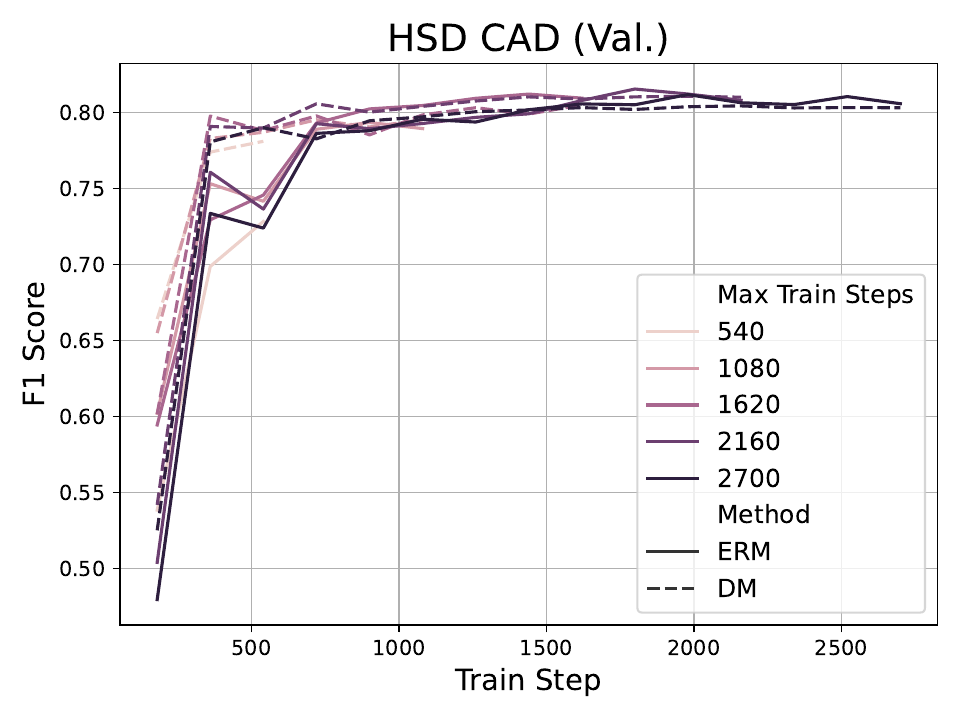}
        \caption{CAD}
        \label{fig:speed_gain_cad}
    \end{subfigure}
    \hfill
    \begin{subfigure}{0.31\textwidth}
        \centering
        \includegraphics[width=\linewidth]{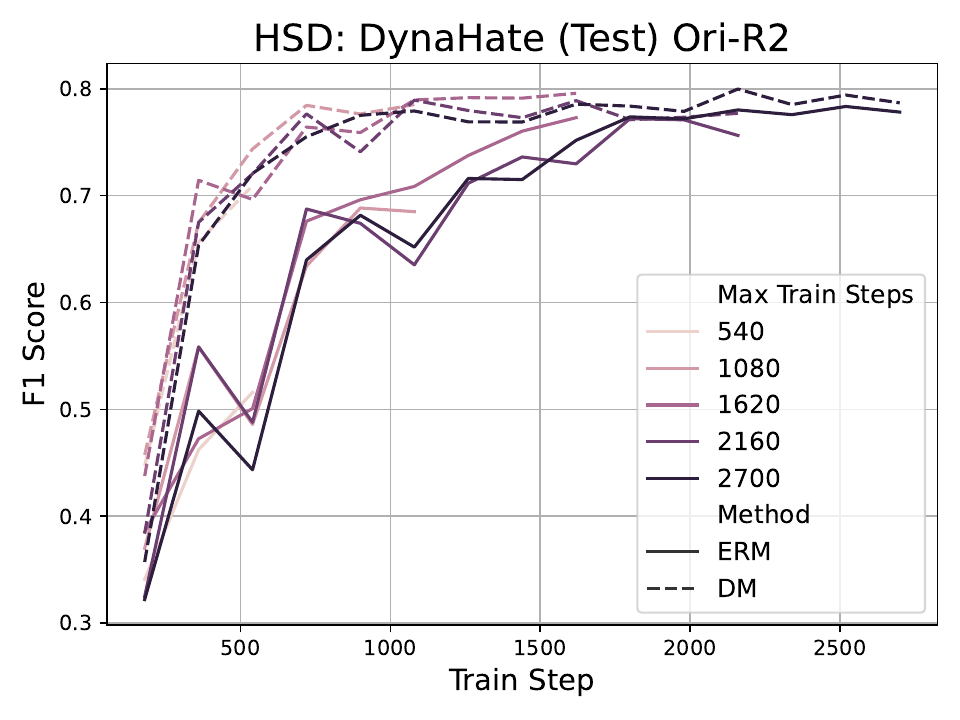}
        \caption{DynaHate}
        \label{fig:speed_gain_dynahate}
    \end{subfigure}
    \caption{
        Figure~\ref{fig:transferability_sizes_instance_level}: 
        Consistency across different sizes of DeBERTaV3 on NLI. 
        The numbers are the percentages (0–1) 
        of ambiguous training instances shared by two models. 
        Training dynamics are transferable across different model sizes: 
        the percentages between models of different sizes 
        are only slightly smaller than those between
        models of different random seeds (shown as superscript).
        Figures~\ref{fig:speed_gain_cad} \& \ref{fig:speed_gain_dynahate}: 
        Performance on HSD when training the main model ($\text{DeBERTaV3}_{\text{Large}}$) 
        using different numbers of training steps. 
        We experimented with different lengths of training (max training steps), 
        and different methods (using ERM and DM). 
        Training with data instances selected by DMs 
        achieves consistently higher training speed than ERM: 
        for datasets on which training with DM achieves 
        either better (OOD datasets, right) or worse (ID datasets, left) performance, 
        models trained with DM outperform ERM with fewer training steps 
        (i.e. the early stage of training, the leftmost part of the x-axis). 
    }
    \label{fig:speed_gain}
\end{figure*}

In this section, we study the transferability of training dynamics in dataset cartography, 
i.e., whether we can use different reference and main models
while maintaining the robustness advantage of the main model.  
Specifically, we study whether training dynamics are transferable across 
different model sizes (\S\ref{subsec:transferability_across_sizes}, 
e.g., from $\text{DeBERTaV3}_{\text{Small}}$ to $\text{DeBERTaV3}_\text{Large}$) 
and pretraining methods (\S\ref{subsec:transferability_across_model_architectures}, 
e.g., from $\text{ELECTRA}_{\text{Large}}$ to $\text{DeBERTaV3}_\text{Large}$).

We focus on these issues for two reasons. 
First, transferability across model sizes enables 
using more efficient (and usually less capable) reference models, 
which (1) can improve training efficiency, and 
(2) is potentially more effective in identifying ambiguous/hard instances,
because they are usually worse at memorizing training instances. 
Second, transferability across pretraining methods 
can help achieve these advantages 
even in cases where more efficient variants of the main pretraining method 
are unsuitable or unavailable for the task.
Moreover, understanding transferability can shed light on data importance: 
If DMs of different reference models consistently identify 
the same subset of training instances as ambiguous, 
it suggests that DMs reveal intrinsic data characteristics. 

We define successful transfers as transfers that produce 
comparable or better OOD performance than ERM. 
Our results show that, with a few exceptions, training dynamics are indeed transferable.
To understand the conditions for successful transfers, 
we analyze the failure cases (\S\ref{subsec:how_efficient_can_we_be}). 
We find that the DMs of reference models that lead to successful transfers typically 
identify a larger subset as easy, which serves as a rough indicator of their capability.
This finding can serve as a guideline for choosing reference models 
without training the main model, 
which is computationally expensive.

\subsection{Transferability Across Model Sizes}\label{subsec:transferability_across_sizes}

In this section, we study 
whether smaller and more efficient models can be used as reference models 
to construct DMs for training larger main models 
(e.g. $\text{DeBERTaV3}_{\text{Small}}$ as the reference model for $\text{DeBERTaV3}_{\text{Large}}$). 
Levering transferability of this type can improve the efficiency of dataset cartography 
by reducing the cost of constructing DMs.

The results for DeBERTaV3 are shown in 
Table~\ref{tab:nli_overall} under section \emph{across different model sizes}  (NLI) and Table \ref{tab:hsd_overall} (HSD, Appendix \ref{app:additional_results}). 
\emph{Training dynamics are transferable across different model sizes}: 
when using $\text{DeBERTaV3}_\text{Large}$ as the main model,  
changing the reference model to either $\text{DeBERTaV3}_\text{Small}$ or $\text{DeBERTaV3}_\text{Base}$ yields 
comparable or even better performance. 
This observation is consistent with our hypothesis that
efficient models are more sensitive to ambiguous and difficult instances. 
As a result, \emph{they can serve as alternative efficient reference models}. 
We also observe that, consistent with \citet{swayamdipta-etal-2020-dataset}, 
ERM achieves better ID performance (in the baselines section of Table \ref{tab:nli_overall}), 
while data cartography performs comparably or better on OOD data. 
Note that hyper-parameters are tuned for ERM performance (see Appendix \ref{app:experimental_setup}), 
\emph{making the training setup more favorable overall to ERM}. 

To investigate transferability of DMs further, 
we analyze whether reference models of different sizes identify similar groups of ambiguous instances. 
Figure~\ref{fig:transferability_sizes_instance_level} shows 
the percentage of ambiguous instances shared by reference models of different sizes and random seeds. 
The percentages shared between different model sizes 
are only slightly smaller than those between the same size but different random seeds, 
providing further evidence for the transferability of DMs.\footnote{
    The expected overlap between two random DMs is 0.33.}

\subsection{Transferability Across Pretraining Methods}
\label{subsec:transferability_across_model_architectures}

We now study the transferability of training dynamics across different pretraining methods.
Successful transfers of this type enable the use of efficient reference models, 
when there is no efficient version of the main model that suits the downstream task.

The results for $\text{DeBERTaV3}_\text{Large}$ as the main model with different reference models
are shown in 
Table~\ref{tab:nli_overall} under section \emph{across different pretraining methods}  (NLI) and 
Table~\ref{tab:hsd_overall} (HSD, Appendix \ref{app:additional_results}).
\emph{Training dynamics are generally transferable across different pretraining methods}: 
$\text{DeBERTaV3}_\text{Large}$ achieves comparable performance 
using DMs constructed by different reference models in most cases. 
However, 
there is one exception:
when using $\text{ELECTRA}_\text{Small}$ as the reference model, 
the performance is clearly worse on the NLI OOD datasets than when using ERM. 
We hypothesize that $\text{ELECTRA}_\text{Small}$ is not strong enough 
for constructing effective DMs. 
We analyze this hypothesis further in \S\ref{subsec:how_efficient_can_we_be} below. 

\subsection{When Do Transfers Fail?}\label{subsec:how_efficient_can_we_be}

\begin{table*}
    \centering
\begin{adjustbox}{max width=0.85\linewidth}
\begin{tabular}{lll|c|ccccc}
    \toprule
    Method & Main Model & Ref. Model & Compute & MultiNLI (Val.) & \multicolumn{3}{c}{AdversarialNLI (Test)} \\
    &  &  &  & - & R1 & R2 & R3 \\
    \midrule
    \multicolumn{8}{c}{ERM Performance: Capabilities of Various Models} \\ 
    \midrule
    \rowcolor[HTML]{C0C0C0} 
    ERM & $\text{TinyBERT}$ & - & 1.45 & ${67.29}_{0.26}$ & ${23.50}_{0.82}$ & ${28.30}_{0.36}$ & ${30.69}_{0.42}$ \\
    \rowcolor[HTML]{C0C0C0} 
    ERM & $\text{ELECTRA}_{\text{Small}}$ & - & 4.61 & ${82.08}_{0.09}$ & ${23.82}_{0.75}$ & ${28.93}_{1.16}$ & ${30.54}_{0.70}$ \\
    ERM & $\text{ELECTRA}_{\text{Base}}$ & - & 36.18 & ${88.47}_{0.20}$ & ${35.45}_{0.24}$ & ${30.95}_{0.42}$ & ${31.27}_{0.68}$ \\
    ERM & $\text{DeBERTaV3}_{\text{Small}}$ & - & 14.47 & ${87.76}_{0.09}$ & ${33.25}_{1.55}$ & ${30.07}_{0.66}$ & ${31.89}_{0.43}$ \\
    ERM & $\text{DeBERTaV3}_{\text{Base}}$ & - & 28.29 & ${90.03}_{0.13}$ & ${43.73}_{0.61}$ & ${33.95}_{0.49}$ & ${33.79}_{1.11}$ \\
    ERM & $\text{BERT}_{\text{Large}}$ & - & 113.49 & ${86.25}_{0.26}$ & ${20.12}_{0.85}$ & ${29.05}_{1.02}$ & ${29.89}_{0.57}$ \\
    ERM & $\text{RoBERTa}_{\text{Large}}$ & - & 116.78 & ${89.86}_{0.09}$ & ${43.85}_{0.64}$ & ${28.55}_{1.07}$ & ${26.00}_{1.04}$ \\
    \midrule
    \multicolumn{8}{c}{When Do Transfers Fail: Using Reference Models with Different Capabilities} \\ 
    \midrule
    \rowcolor[HTML]{C0C0C0} 
    DM & $\text{DeBERTaV3}_{\text{Large}}$ & $\text{TinyBERT}$ & 101.45 & ${89.26}_{0.16}$ & ${42.47}_{0.87}$ & ${34.38}_{0.66}$ & ${33.56}_{0.52}$ \\
    \rowcolor[HTML]{C0C0C0} 
    DM & $\text{DeBERTaV3}_{\text{Large}}$ & $\text{ELECTRA}_{\text{Small}}$ & 104.61 & ${90.84}_{0.03}$ & ${50.68}_{1.55}$ & ${39.92}_{0.55}$ & ${37.10}_{0.87}$ \\
    DM & $\text{DeBERTaV3}_{\text{Large}}$ & $\text{ELECTRA}_{\text{Base}}$ & 136.18 & ${90.28}_{0.19}$ & ${60.75}_{0.98}$ & ${47.30}_{0.68}$ & ${42.92}_{0.87}$ \\
    DM & $\text{DeBERTaV3}_{\text{Large}}$ & $\text{DeBERTaV3}_{\text{Small}}$ & 114.47 & ${90.77}_{0.16}$ & ${60.10}_{1.46}$ & ${46.23}_{0.41}$ & ${41.46}_{0.94}$ \\
    DM & $\text{DeBERTaV3}_{\text{Large}}$ & $\text{DeBERTaV3}_{\text{Base}}$ & 128.29 & ${90.64}_{0.22}$ & ${59.67}_{1.10}$ & ${45.88}_{1.01}$ & ${43.00}_{1.44}$ \\
    DM & $\text{DeBERTaV3}_{\text{Large}}$ & $\text{BERT}_{\text{Large}}$ & 213.49 & ${89.37}_{0.15}$ & ${62.00}_{0.92}$ & ${48.60}_{0.65}$ & ${44.73}_{0.48}$ \\
    DM & $\text{DeBERTaV3}_{\text{Large}}$ & $\text{RoBERTa}_{\text{Large}}$ & 216.78 & ${90.26}_{0.06}$ & ${61.02}_{1.03}$ & ${46.85}_{0.54}$ & ${42.65}_{0.83}$ \\
    \midrule
    \rowcolor[HTML]{C0C0C0} 
    DM & $\text{ELECTRA}_{\text{Large}}$ & $\text{TinyBERT}$ & 111.64 & ${88.30}_{0.22}$ & ${35.37}_{0.31}$ & ${32.60}_{1.25}$ & ${30.78}_{0.64}$ \\
    \rowcolor[HTML]{C0C0C0} 
    DM & $\text{ELECTRA}_{\text{Large}}$ & $\text{ELECTRA}_{\text{Small}}$ & 114.80 & ${90.35}_{0.19}$ & ${45.38}_{0.67}$ & ${34.95}_{0.90}$ & ${32.58}_{1.16}$ \\
    DM & $\text{ELECTRA}_{\text{Large}}$ & $\text{ELECTRA}_{\text{Base}}$ & 146.38 & ${89.96}_{0.20}$ & ${55.40}_{0.28}$ & ${42.00}_{0.42}$ & ${36.71}_{0.65}$ \\
    DM & $\text{ELECTRA}_{\text{Large}}$ & $\text{DeBERTaV3}_{\text{Small}}$ & 124.67 & ${90.27}_{0.13}$ & ${54.20}_{0.57}$ & ${40.00}_{1.13}$ & ${36.88}_{0.06}$ \\
    DM & $\text{ELECTRA}_{\text{Large}}$ & $\text{DeBERTaV3}_{\text{Base}}$ & 138.49 & ${89.94}_{0.14}$ & ${53.63}_{0.80}$ & ${41.33}_{0.51}$ & ${36.33}_{0.25}$ \\
    DM & $\text{ELECTRA}_{\text{Large}}$ & $\text{BERT}_{\text{Large}}$ & 223.68 & ${88.83}_{0.12}$ & ${57.10}_{0.79}$ & ${43.27}_{0.84}$ & ${38.72}_{0.31}$ \\
    DM & $\text{ELECTRA}_{\text{Large}}$ & $\text{RoBERTa}_{\text{Large}}$ & 226.97 & ${89.62}_{0.13}$ & ${54.20}_{0.57}$ & ${42.55}_{0.78}$ & ${37.29}_{0.65}$ \\
    \bottomrule
\end{tabular}
\end{adjustbox}
\caption{
    We use reference models with different capabilities 
    (measured by their ERM accuracy) to construct DMs,  
    and use them to train main models.
    The gray-shaded rows are (1) the main models in unsuccessful transfers 
    and (2) their corresponding reference models. 
    Successful transfer requires the reference model to be reasonably strong:
    reference models with clearly worse ID performance lead to 
    degraded OOD performance for the main models. 
}
\label{tab:how_efficient_can_we_be}
\end{table*}

We have shown that training dynamics are usually transferable 
across different model sizes and pretraining methods.  
We now study the conditions for successful transfers, by zooming in on two questions: 
1) Can we use efficient but weak models as reference models? 
and 2) What are the differences between effective and ineffective reference models?
Answers to these questions can guide the selection of efficient yet effective reference models.

\paragraph{Can we use efficient but weak models as reference models?} 

To answer this question, we compare the performance of a wide range of methods, 
see Table~\ref{tab:how_efficient_can_we_be} (NLI) 
and Table~\ref{tab:how_efficient_can_we_be_hsd} (HSD, in Appendix~\ref{app:additional_results}). 
We include models from three categories: 
First, a wide range of seven models trained using ERM 
(section \emph{ERM Performance}): 
TinyBERT, the small and base versions of DeBERTaV3 and ELECTRA, 
$\text{RoBERTa}_{\text{Large}}$, and $\text{BERT}_{\text{Large}}$. 
Second, we use these ERM models as reference models to construct DMs, 
and use these DMs to fine-tune $\text{DeBERTaV3}_\text{Large}$
(section \emph{When Do Transfers Fail}, first half). 
By using reference models with different sizes and pretraining methods to fine-tune the same main model, 
we can inspect the impact of reference model capability on transferability. 
Third, we also include the results of using $\text{ELECTRA}_\text{Large}$ as the main model
(section \emph{When Do Transfers Fail}, second half).
By comparing results from different main models, 
we can better understand whether successful transfers originate from 
the compatibility between reference and main models, 
or the capability of the reference model itself.
We make three observations.

First, weak reference models with poor ID performance, i.e., TinyBERT and $\text{ELECTRA}_\text{Small}$, 
lead to unsuccessful transfers. 
However, the reference models do not need to be as capable as the main model: 
slightly weaker reference models could be more useful, 
indicated by the strong OOD performance when using $\mathrm{BERT}_{\mathrm{Large}}$ as the reference model. 
Moreover, in these unsuccessful transfers, 
the main model OOD performance correlates with the reference model ID performance. 
For example, on NLI, TinyBERT has a lower ID performance than $\text{ELECTRA}_\text{Small}$, 
and both main models $\text{DeBERTaV3}_\text{Large}$ and $\text{ELECTRA}_\text{Large}$ 
have a lower OOD performance when using TinyBERT as the reference model 
compared to when using $\text{ELECTRA}_\text{Small}$ as the reference model. 

Second, whether a transfer is successful or not depends mostly on the reference model, 
rather than the compatibility between the reference and the main models: 
The transfers from TinyBERT and $\text{ELECTRA}_\text{Small}$ to both main models are unsuccessful, 
and the transfers from other reference models to both main models are successful.\footnote{
    We also include results using $\text{DeBERTaV3}_\text{Base}$ as the main model, 
    and TinyBERT and $\text{DeBERTaV3}_\text{Small}$ as reference models on HSD 
    in Table~\ref{tab:how_efficient_can_we_be_hsd}. 
    Transfers from $\text{DeBERTaV3}_\text{Small}$ substantially outperform ERM, 
    while those from TinyBERT underperform ERM,  
    further suggesting that the reference model is decisive for successful transfers.
}

Third, interestingly, ID performance in all transfers remains high. 
In other words, ineffective reference models only impact the OOD performance of the main models. 
For instance, the transfer from $\text{ELECTRA}_\text{Small}$ to $\text{DeBERTaV3}_\text{Large}$ 
yields the best accuracy on MultiNLI despite its OOD performance being relatively poor. 
We suspect that
weak models with poor ID performance often identify easy training instances as ambiguous data. 
While easy instances can be sufficient for obtaining satisfactory ID performance, 
they are insufficient for good OOD performance \citep{swayamdipta-etal-2020-dataset}. 
We discuss this in detail below.

\paragraph{What are the differences between effective and ineffective reference models?}

\begin{figure}[ht]
    \centering
    \includegraphics[width=.4\textwidth]{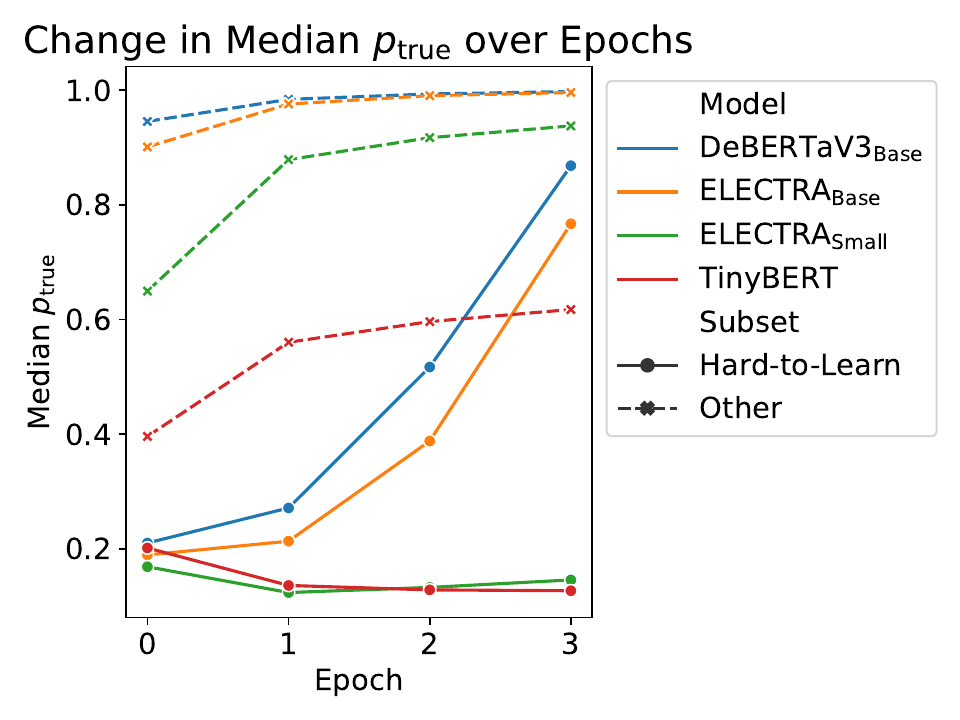}
    \caption{
        Change of median $p_{\text{true}}$: 
        ineffective reference models ($\text{ELECTRA}_{\text{Small}}$ and TinyBERT) are unable to fit difficult training instances, 
        making easy instances being identified as ambiguous. 
    }
    \label{fig:change_median}
\end{figure}

To answer this question, 
we consider the possible differences between a weak (and ineffective) reference model 
and a reasonably strong reference model, 
in terms of categorizing training data into ambiguous, hard-to-learn, and easy subsets. 
Also, we assume that instances in our training set exhibit varying levels of difficulty (i.e. simple to difficult). 

Assume we have a weak reference model that 
can learn simpler training instances but cannot learn the more difficult ones. 
This weak reference model will therefore assign increasing $p_{\text{true}}$ 
to simple training instances across different epochs, 
while keeping $p_{\text{true}}$ for difficult training instances 
around the values expected in a random guessing scenario.
Consequently, $p_{\text{true}}$ will exhibit high standard deviations on simpler training instances, 
which will then be identified as ambiguous data; 
while more difficult training instances will have 
consistent lower mean values and thus low standard deviations for $p_{\text{true}}$, 
and therefore be identified as hard-to-learn data. 
In contrast, a reasonably capable reference model 
can learn simpler training instances during the early stage of training 
(even before their first occurrence in the training batch, 
i.e., their correct predictions are learned from other training instances). 
Therefore, these instances will have 
both high mean values and low standard deviations for $p_{\text{true}}$. 
Meanwhile, 
$p_{\text{true}}$ for difficult instances will gradually increase across epochs, 
making these instances yield 
relatively low mean values and high standard deviations for $p_{\text{true}}$. 
As a result, these instances will
be identified as both ambiguous and hard-to-learn (i.e. we expect a large overlap in these subsets).  
Because we select a fixed percentage $q\%$ of instances as ambiguous or hard-to-learn, 
this larger overlap means a larger easy subset too.

We now validate our reasoning. 
Given a reference model, 
we first split the training instances into two subsets based on mean $p_{\text{true}}$: 
hard-to-learn ($10\%$ of training instances) and other (the remaining $90\%$). 
We use a lower $q\%$ to make the difference clearer. 
Then, for each subset, 
we calculate the median $p_{\text{true}}$ in each epoch. 
We use median values because they are robust statistics of the central tendency. 
Figure~\ref{fig:change_median} shows our results on MultiNLI, 
using two effective ($\text{DeBERTaV3}_{\text{Base}}$ and  $\text{ELECTRA}_{\text{Base}}$) 
and two ineffective ($\text{ELECTRA}_{\text{Small}}$ and TinyBERT) reference models. 
We only include four models to make the figure clearer --- 
we observe similar trends for other models: 
With effective reference models, 
hard-to-learn instances are gradually learned during training, 
while other instances already have high $p_\text{true}$ values from the first epoch. 
In contrast, with ineffective reference models, 
hard-to-learn data instances are not learned at all, 
suggested by their close-to-zero $p_\text{true}$ over different epochs; 
while other instances are gradually fitted, indicated by their increasing values. 

To further validate our reasoning, 
we compute the percentages of training instances identified as easy by different reference models 
(Table~\ref{tab:easy_data_ratio} in Appendix~\ref{app:additional_results}): 
We report results with  $q \in \{10\%, 25\%, 33\%, 50\%\}$. %
Less effective reference models indeed identify fewer data points as easy. 
For example, on NLI with $q\%=50\%$, 
TinyBERT identifies less than 20.0\% of the instances as easy,
compared to 46.65\% by $\text{DeBERTaV3}_{\text{Base}}$.
Furthermore, 
the overlap between hard-to-learn and ambiguous instances in successful transfers 
is usually high. 
For example, with $q\%=50\%$, 
all effective reference models identify more than 46\% as easy training instances 
(the maximum is 50\%, when the ambiguous and hard-to-learn subsets overlap perfectly).

\section{Training Speed Gains From Data Selection}\label{sec:efficiency_gain}

Our results from \S\ref{sec:transferability_of_training_dynamics} suggest that 
the efficiency of dataset cartography can be improved with more efficient reference models. 
However, if we train the main models for the same number of steps as ERM, 
the computational cost of the full pipeline is still higher than conventional fine-tuning (i.e., ERM), 
because of the extra reference model training cost. 

Previous studies have shown that 
training with a careful selection of ``informative" training instances 
can lead to higher training speed than ERM (i.e. achieving the same performance with fewer training steps), 
despite the computational cost of selecting these instances being notoriously high
\citep{sorscher2022beyond,NEURIPS2020_1e14bfe2,paul2021deep,toneva2018an}. 
Motivated by this,  
we now study whether we can achieve similar learning speed gains 
by training on instances selected by DMs.
If such gains exist, 
we can further improve the efficiency of dataset cartography by training with fewer steps. 

We show the performance of $\text{DeBERTaV3}_{\text{Large}}$ on HSD, 
fine-tuned using the full training set (ERM) and using only the instances selected by DMs, 
across different training durations (i.e. max training steps) in Figure~\ref{fig:speed_gain}. 
Results on other datasets are in Appendix~\ref{app:additional_results}, 
where we observe similar trends.
We make three observations. 
First, when the number of training steps is reduced, 
models fine-tuned with dataset cartography clearly outperform those fine-tuned with ERM, 
even on ID validation data where ERM achieves better performance with more training steps. 
Second, 
dataset cartography often yields better results with fewer training steps than with more, 
suggesting a tendency of overfitting at the late training stage. 
Third, the efficiency gains hold consistently across different training durations. 
Our observations indicate that 
\emph{training with data instances selected by DMs 
consistently achieves higher training speed} than ERM.

\section{Our Approach: FTFT}\label{sec:ftft}

Building on our insights presented in the previous sections, 
we propose a novel fine-tuning approach based on dataset cartography~\citep{swayamdipta-etal-2020-dataset}, 
\textbf{Fine-Tuning by transFerring Training dynamics (FTFT)}.
Compared with the original dataset cartography approach, 
FTFT integrates two crucial improvements: 
(1) using more efficient reference models, 
which not only result in comparable DMs but also enhance efficiency, 
and (2) implementing aggressive early stopping during the main model training. 
Such aggressive early stopping is possible because models
trained with data instances selected using DMs
already achieve strong performance with substantially fewer training steps
(\S\ref{sec:efficiency_gain}). 
In summary, FTFT consists of three steps: 
(1) train an efficient reference model on the full dataset, 
(2) compute the ambiguity of each instance 
based on the standard deviation of the correct class probability across epochs, and 
(3) select the top $q\%$ most ambiguous instances to retrain a larger main model, 
while applying aggressive early stopping.

We show our results 
on NLI in Table~\ref{tab:nli_overall} (FTFT) and
on HSD in Table~\ref{tab:hsd_overall} (Appendix~\ref{app:additional_results}). 
We use an early stopping patience $k=2$
(i.e. the number of checkpoints without improvement before stopping), 
based on the average dev set performance on the OOD datasets. 
We use a relatively small $k$, as stopping the training earlier will lead to higher efficiency.
We also show the computational cost of each method (Compute), 
in which we have taken the training of 
both the reference model and the main model into account, 
including the $k$ extra checkpoints after the best performance is achieved.

We have three key observations. 
First, FTFT achieves consistent robustness improvements over ERM, 
indicated by its strong performance on most OOD datasets
(note again that the hyper-parameters are optimized for ERM). 
We also include the results of ERM with early stopping (ERM(ES)) as a baseline 
to illustrate that our aggressive early stopping strategy is only effective 
when training with data instances selected by DMs: 
ERM(ES) achieves worse performance than ERM on all OOD datasets.
Second, FTFT substantially improves the training efficiency of models. 
For example, on NLI, training $\text{DeBERTaV3}_{\text{Large}}$ using 
FTFT and $\text{DeBERTaV3}_{\text{Small}}$ as the reference model 
only costs 51.57\% of ERM's training cost,
which is equivalent to 25.78\% of the training cost 
of the original dataset cartography approach. 
Third, our early stopping strategy is effectively aggressive. 
Specifically, all FTFT results in Table~\ref{tab:nli_overall} 
are achieved \textbf{using less than 1/3 of the optimal training steps needed for ERM}.

\section{Conclusion}

Fine-tuned PLMs have shown to be vulnerable to OOD input. 
Although dataset cartography can improve model robustness 
\citep{swayamdipta-etal-2020-dataset,sar-shalom-schwartz-2023-curating},
it is computationally expensive. 
In this paper, 
we have presented FTFT, a novel approach for fine-tuning PLMs 
which yields both better efficiency and better robustness over ERM 
(\S\ref{sec:ftft}). 
FTFT is built on dataset cartography, based on two observations: 
(1) reference model training dynamics are highly transferable across different 
model sizes (\S\ref{subsec:transferability_across_sizes}) 
and pretraining methods~(\S\ref{subsec:transferability_across_model_architectures}), and 
(2) main models trained using data instances selected using DMs 
learn faster (i.e. they have substantially better performance with reduced number of training steps). 
We believe that FTFT will be an important tool for future researchers and practitioners 
to perform efficient model fine-tuning, 
especially in situations where robustness is essential. 

\section{Limitations}
We identify three limitations from this work. 
These limitations open up promising directions for future research. 
First, we have observed that 
effective reference models identify more instances as easy. 
More controlled experiments are needed 
to build a detailed protocol for choosing efficient but strong enough reference models. 
Our results provide a good foundation for such further work.
Second, we have empirically demonstrated the transferability 
of training dynamics to select training instances. 
Future studies are needed to build the theoretical foundations of 
both data cartography itself and the feasibility of such transfers. 
Third, we have only developed FTFT for classification tasks. 
Future studies should extend FTFT to generation tasks, 
e.g., instruction following and language modeling.

\section*{Acknowledgments}
We thank the anonymous reviewers for their helpful feedback.
This work used the Dutch national e-infrastructure with the support of the
SURF Cooperative using grants no. EINF-10999 and EINF-5693.

\bibliography{custom}

\clearpage
\appendix

\section{Training Specifications}

\subsection{Experimental Setup}\label{app:experimental_setup}

\paragraph{Optimization}
For training all models, 
we use AdamW~\citep{loshchilov2018decoupled} as the optimizer with a batch size of 32.  
We also use a linear learning rate scheduler with 10\% warmup.
For fine-tuning the small and base versions of 
both DeBERTaV3 and ELECTRA, as well as TinyBERT, RoBERTa, and BERT, 
we use a learning rate of 2e-5, 
following the suggestions from the original papers. 
We use a smaller learning rate of 1e-5 for $\mathrm{DeBERTa}_{\mathrm{Large}}$ and $\mathrm{ELECTRA}_{\mathrm{Large}}$, 
because training larger models with lower learning rates are observed to be more stable 
\citep{mosbach2021on,du-nguyen-2023-measuring}. 
However, a few failed training runs 
(i.e., runs where the training fails to converge and 
where the resulting model performs worse than the majority class baseline~\citep{mosbach2021on}) 
still occurred during the training of $\mathrm{ELECTRA}_{\mathrm{Large}}$. 
We excluded these runs from our results.

\paragraph{Number of Training Steps and Checkpointing}
Regarding the number of training steps, 
we perform a grid search for 
both $\mathrm{DeBERTa}_{\mathrm{Large}}$ and $\mathrm{ELECTRA}_{\mathrm{Large}}$ ERM models, 
spanning from one to five epochs, 
according to their average performance on the validation set of AdversarialNLI (NLI) and DynaHate (HSD). 
On NLI, the optimal length of training is four epochs (49088 steps) and five epochs (61360 steps) 
for $\mathrm{DeBERTa}_{\mathrm{Large}}$ and $\mathrm{ELECTRA}_{\mathrm{Large}}$. 
On HSD, the optimal length of training is three epochs (1620 steps) 
for both $\mathrm{DeBERTa}_{\mathrm{Large}}$ and $\mathrm{ELECTRA}_{\mathrm{Large}}$. 
For other PLMs, because we only use them as reference models, 
we do not perform this grid search, and use 61360 steps and 1620 steps for NLI and HSD. 
We perform checkpointing every 4090 steps and 180 steps on NLI and HSD, 
which is approximately the length of one epoch using the 33\% selected data instances from DMs. 

\paragraph{Software and Hardware}
We use Python 3.9 and PyTorch 2.0 for all experiments. 
We also use HuggingFace Transformers 4.32 \citep{wolf-etal-2020-transformers}, 
Accelerate 0.22, and Datasets 2.14 \citep{lhoest-etal-2021-datasets}.
All experiments are performed on one NVIDIA A100 GPU.
Training all models (including hyper-parameter search) takes approximately 16 GPU days. 

\subsection{Comparison of Training Costs}\label{app:training_costs}

\begin{table}[h]
    \centering
    \begin{tabular}{lll}
    \toprule
        Model & Param Size & Compute \\
        \midrule
        $\mathrm{DeBERTaV3}_{\mathrm{Small}}$ & 44.00M & 14.47 \\
        $\mathrm{DeBERTaV3}_{\mathrm{Base}}$ & 86.00M & 28.29 \\
        $\mathrm{DeBERTaV3}_{\mathrm{Large}}$ & 304.00M & 100.00 \\
        $\mathrm{ELECTRA}_{\mathrm{Small}}$ & 14.00M & 4.61 \\
        $\mathrm{ELECTRA}_{\mathrm{Base}}$ & 110.00M & 36.18 \\
        $\mathrm{ELECTRA}_{\mathrm{Large}}$ & 335.00M & 110.20 \\
        $\mathrm{BERT}_{\mathrm{Large}}$ & 345.00M & 113.49 \\
        $\mathrm{RoBERTa}_{\mathrm{Large}}$ & 355.00M & 116.78 \\
        $\mathrm{TinyBERT}$ & 4.40M & 1.45 \\
        \bottomrule
    \end{tabular}
    \caption{Model sizes and their estimated relative training cost in terms of FLOPs, estimated by following~\citet{kaplan2020scaling}.}
    \label{tab:train_compute}
\end{table}

We estimate the training FLOPs by their models sizes (i.e. the number of parameters), 
following~\citet{kaplan2020scaling} ($C\sim6ND$, where 
$C$ is the number of FLOPs, $N$ is the number of parameters, and $D$ is the dataset size). 
The results are in Table~\ref{tab:train_compute}. 
We prefer theoretical estimation over practical measurement, because 
(1) the results are less influenced by noises, 
(2) it considers both forward and backward propagation, and 
(3) it has been shown effective and widely used~\citep{kaplan2020scaling,hoffmann2022training}. 
In practice, 
we also measured the actual FLOPs usage in a forward run using \citet{ptflops}, 
and the results are very close to our estimation using model sizes.

\section{Additional Results}\label{app:additional_results}

\begin{table*}[]
\centering
\begin{adjustbox}{max width=0.9\linewidth}
    \centering
    \begin{tabular}{lcccc|cccc}
    \toprule
    &\multicolumn{4}{c}{NLI: MultiNLI} & \multicolumn{4}{c}{HSD: CAD} \\
    & 10\% & 25\% & 33\% & 50\% & 10\% & 25\% & 33\% & 50\% \\
    Model &  &  &  &  &  &  &  &  \\
    \midrule
    \rowcolor[HTML]{C0C0C0} 
    $\mathrm{TinyBERT}$ & 80.01\% & 51.27\% & 38.34\% & 19.60\% & 81.89\% & 65.48\% & 58.20\% & 41.73\% \\
    \rowcolor[HTML]{C0C0C0} 
    $\mathrm{ELECTRA}_{\mathrm{Small}}$ & 80.10\% & 55.72\% & 47.40\% & 35.50\% & 82.97\% & 69.70\% & 61.92\% & 41.51\% \\
    $\mathrm{ELECTRA}_{\mathrm{Base}}$ & 84.57\% & 69.00\% & 62.84\% & 46.65\% & 86.10\% & 72.07\% & 63.41\% & 44.54\% \\
    $\mathrm{DeBERTaV3}_{\mathrm{Small}}$ & 82.80\% & 67.54\% & 61.72\% & 46.15\% & 85.44\% & 70.82\% & 62.83\% & 46.54\% \\
    $\mathrm{DeBERTaV3}_{\mathrm{Base}}$ & 85.08\% & 71.58\% & 64.06\% & 46.86\% & 85.49\% & 72.30\% & 63.64\% & 46.14\% \\
    $\mathrm{BERT}_{\mathrm{Large}}$ & 85.21\% & 70.08\% & 63.05\% & 47.21\% & 89.25\% & 72.38\% & 64.15\% & 47.07\% \\
    $\mathrm{RoBERTa}_{\mathrm{Large}}$ & 85.67\% & 71.43\% & 64.24\% & 47.20\% & 88.57\% & 73.21\% & 64.84\% & 47.25\% \\
    \bottomrule
    \end{tabular}
    \end{adjustbox}
    \caption{
        Ratio of easy, i.e. neither ambiguous or hard-to-learn data, under different DM thresholds.
        Effective reference models identify larger easy data subsets,
        compared with the less effective reference models $\mathrm{TinyBERT}$ and $\mathrm{ELECTRA}_{\mathrm{Small}}$
        This also implies that the hard-to-learn subsets and ambiguous subsets 
        identified by effective reference models 
        are very close to each other. 
    }
    \label{tab:easy_data_ratio}
\end{table*}

\begin{figure*}
    \centering
    \begin{subfigure}{0.31\textwidth}
    \includegraphics[width=\linewidth]{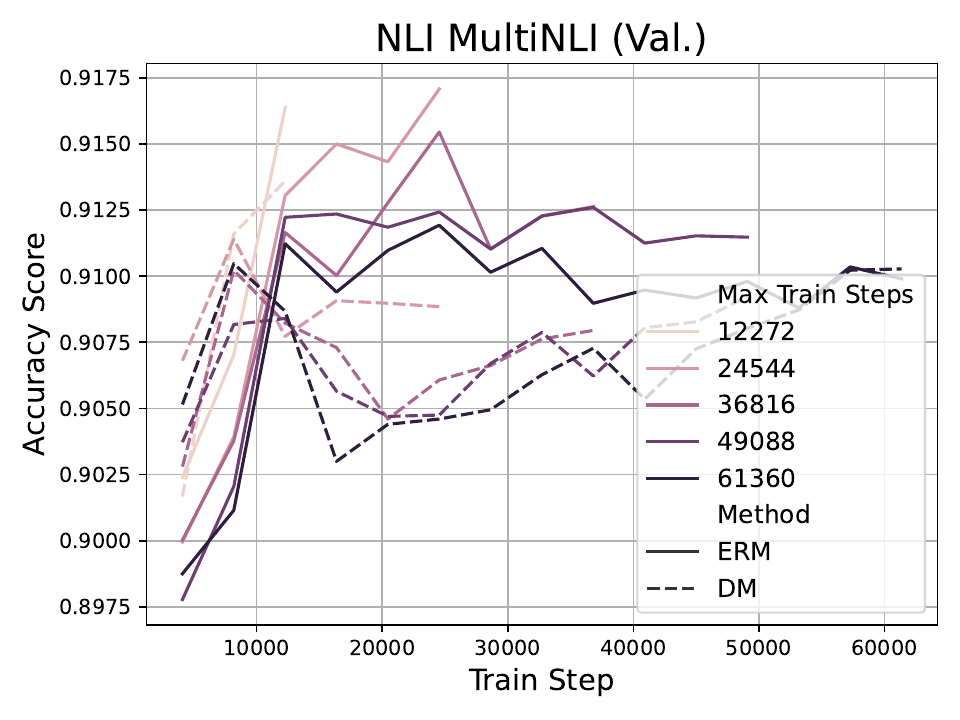}
    \end{subfigure}
    \begin{subfigure}{0.31\textwidth}
    \includegraphics[width=\linewidth]{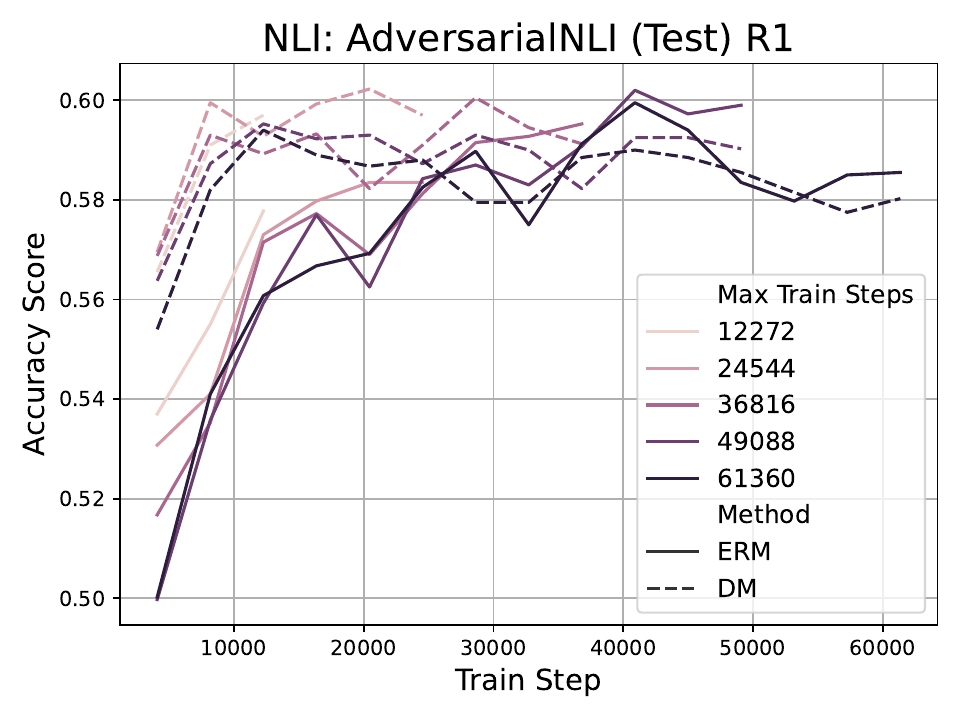}
    \end{subfigure}
    \begin{subfigure}{0.31\textwidth}
    \includegraphics[width=\linewidth]{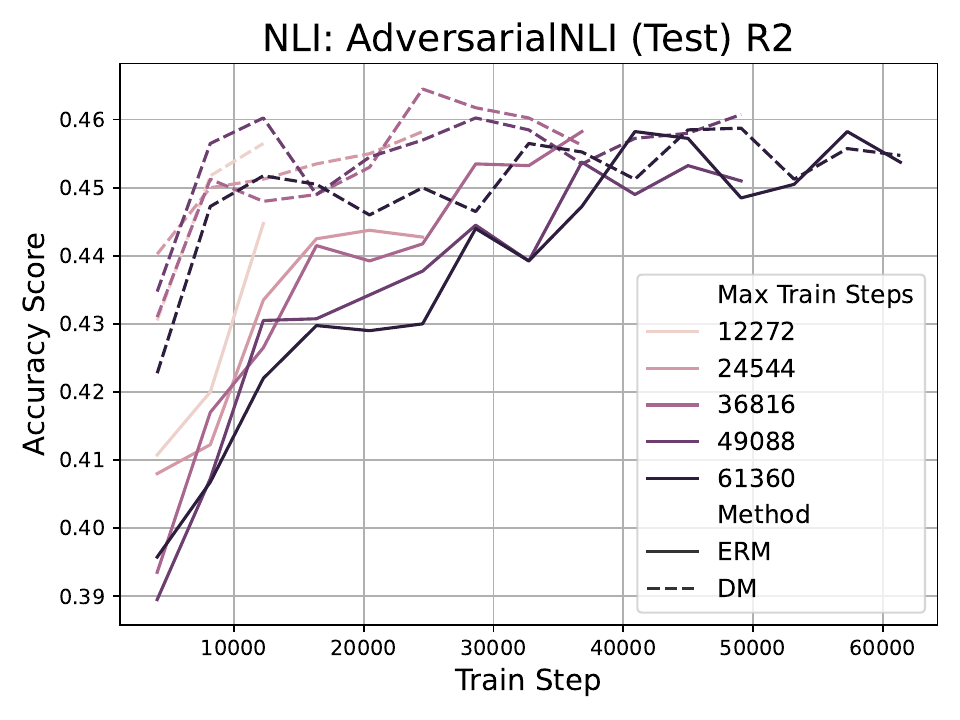}
    \end{subfigure}
    \begin{subfigure}{0.31\textwidth}
    \includegraphics[width=\linewidth]{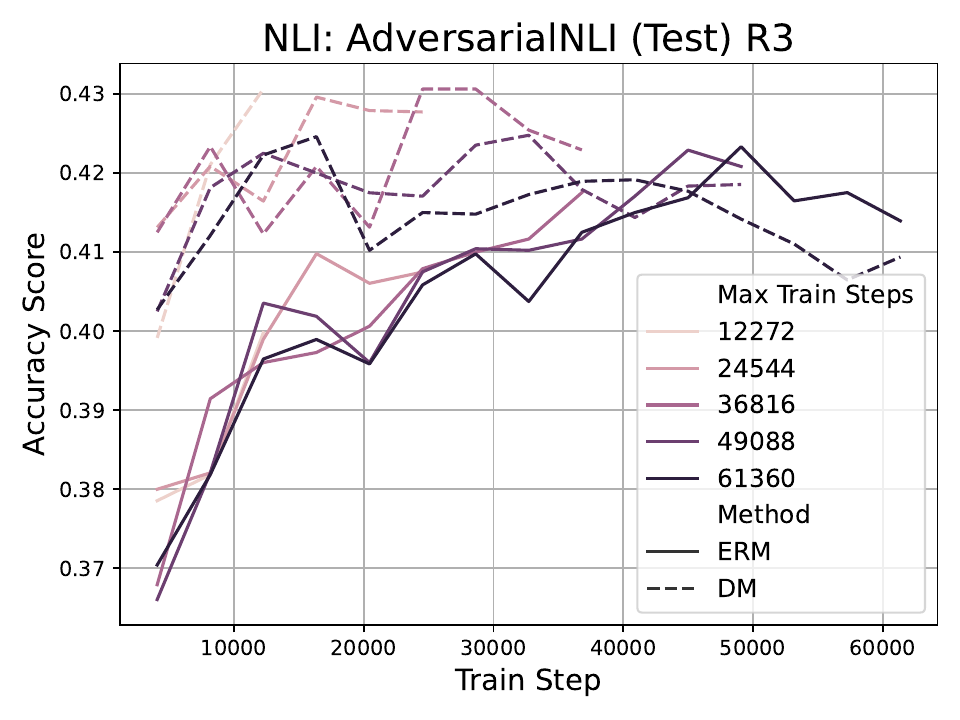}
    \end{subfigure}
    \begin{subfigure}{0.31\textwidth}
    \includegraphics[width=\linewidth]{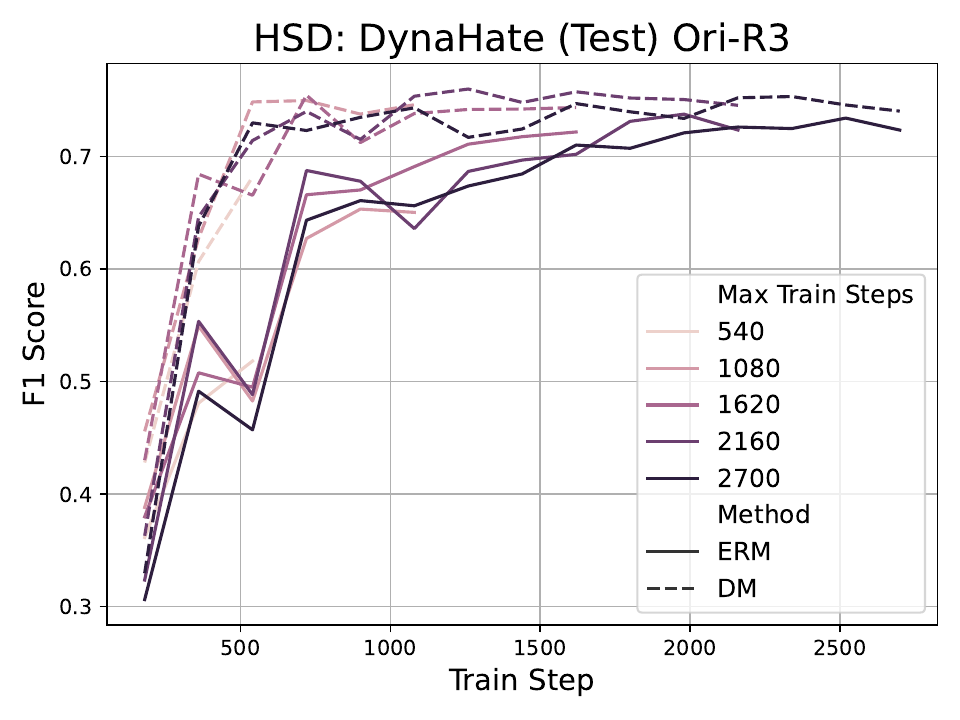}
    \end{subfigure}
    \begin{subfigure}{0.31\textwidth}
    \includegraphics[width=\linewidth]{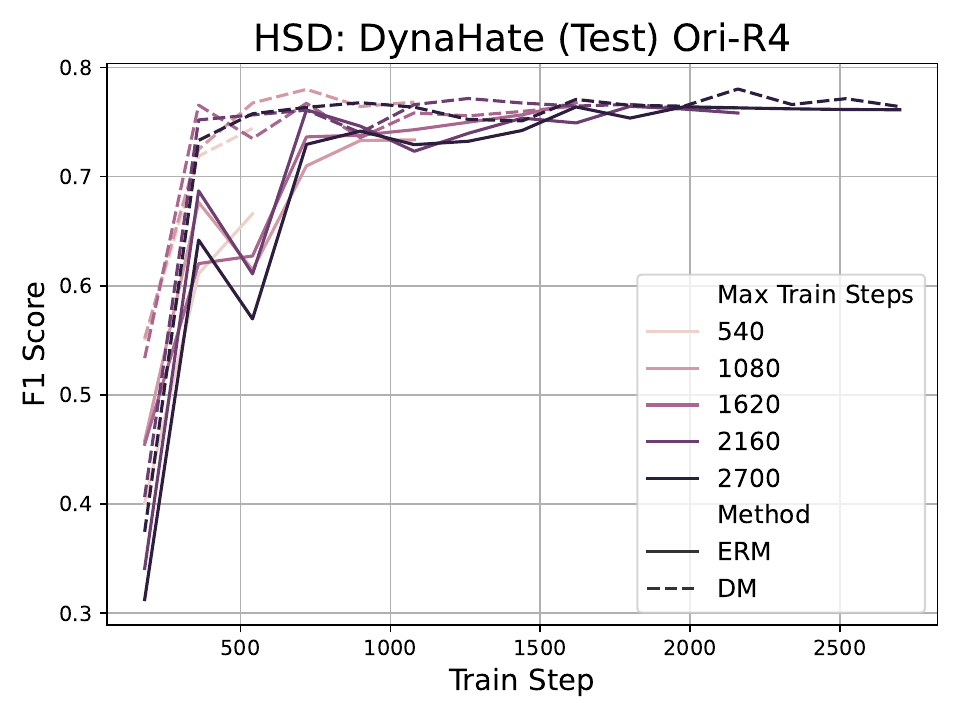}
    \end{subfigure}
    \begin{subfigure}{0.31\textwidth}
    \includegraphics[width=\linewidth]{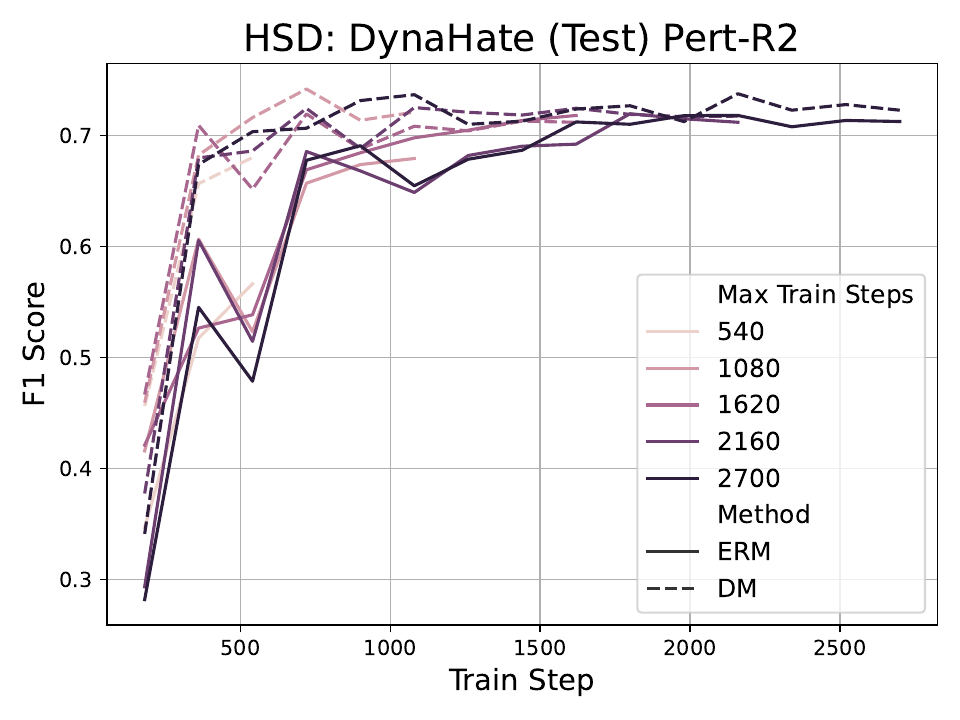}
    \end{subfigure}
    \begin{subfigure}{0.31\textwidth}
    \includegraphics[width=\linewidth]{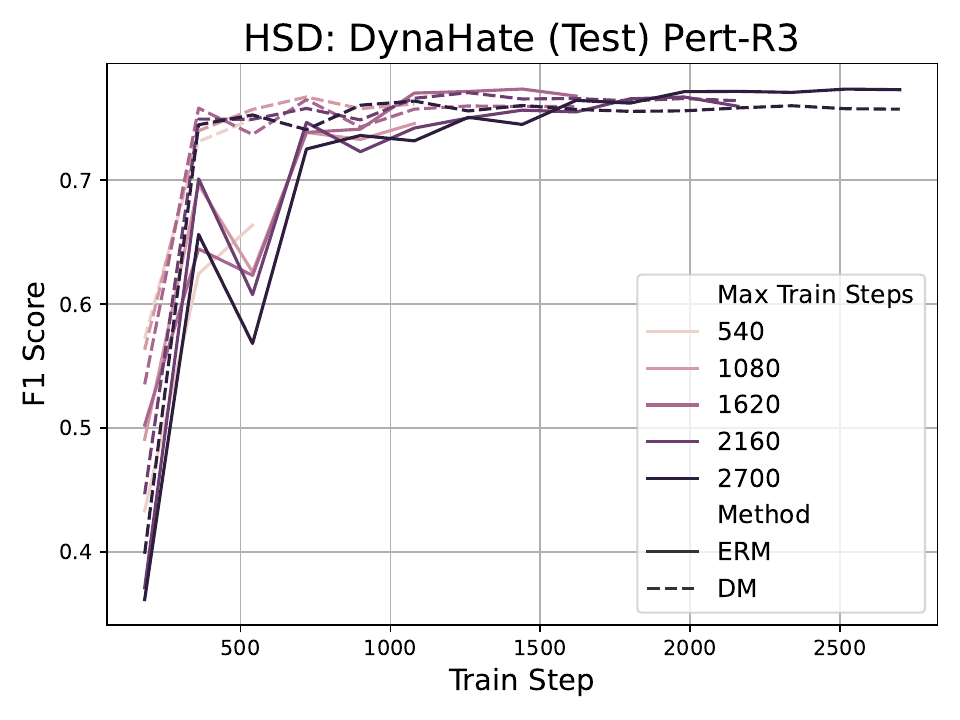}
    \end{subfigure}
    \begin{subfigure}{0.31\textwidth}
    \includegraphics[width=\linewidth]{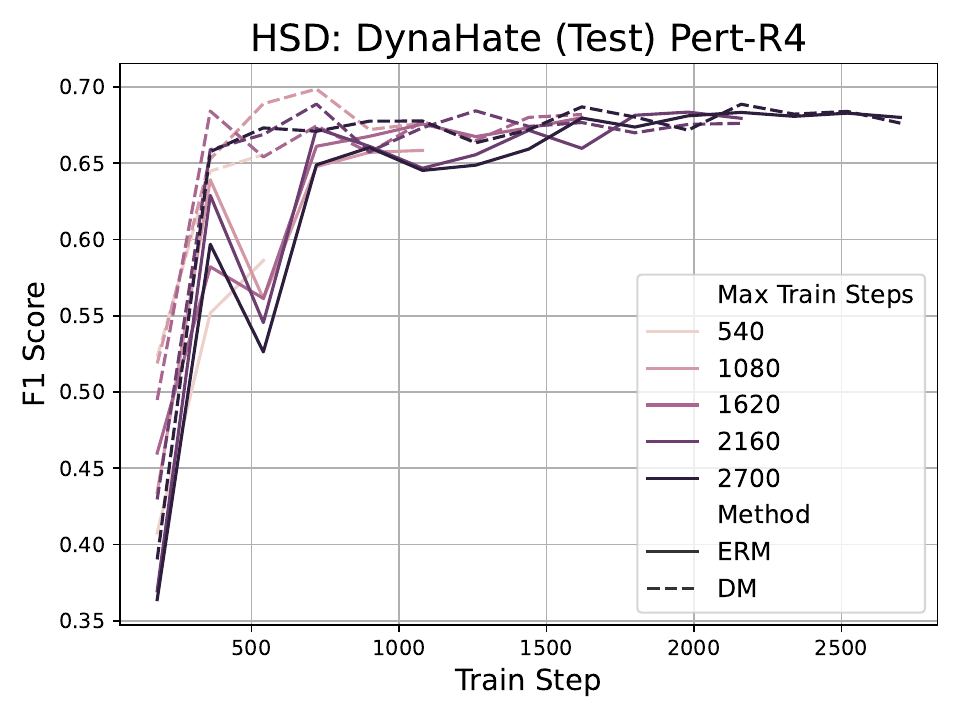}
    \end{subfigure}
    \caption{
        Performance when training the main model ($\text{DeBERTaV3}_{\text{Large}}$) 
        using different numbers of training steps across different checkpoints. 
        We experimented with different lengths of training (max training steps), 
        and different methods (using ERM and DM). 
        Training with data instances selected by DMs 
        achieves consistent higher training speed than ERM: 
        for datasets on which training with DM achieves 
        either better or worse performance, 
        models trained with DM outperform ERM with reduced training steps 
        (i.e. the early stage of training, the leftmost part of the x-axis). 
    }
    \label{fig:speed_gain_more}
\end{figure*}

\clearpage
\begin{sidewaystable*}
\centering
\begin{tabular}{llllcccccccc}
    \toprule
    Method & Main Model & Ref. Model & Compute & CAD (Val.) & \multicolumn{6}{c}{DynaHate (Test)} \\
    &  &  &  & - & Ori-R2 & Ori-R3 & Ori-R4 & Pert-R2 & Pert-R3 & Pert-R4 \\
    \midrule
    \multicolumn{11}{c}{Baselines: ERM, ERM with Early Stopping, and Random DM} \\ 
    \midrule
    ERM & $\text{DeBERTaV3}_{\text{Small}}$ & - & 14.47 & ${75.92}_{1.33}$ & ${50.63}_{2.65}$ & ${52.31}_{2.60}$ & ${61.37}_{1.44}$ & ${57.98}_{2.42}$ & ${65.42}_{0.92}$ & ${61.42}_{0.50}$ \\
    ERM & $\text{DeBERTaV3}_{\text{Base}}$ & - & 28.29 & ${77.70}_{1.11}$ & ${55.97}_{0.75}$ & ${58.88}_{2.84}$ & ${63.79}_{1.84}$ & ${59.76}_{0.94}$ & ${66.63}_{1.53}$ & ${61.07}_{0.95}$ \\
    ERM* & $\text{DeBERTaV3}_{\text{Large}}$ & - & 100.00 & ${80.95}_{0.63}$ & ${77.30}_{1.45}$ & ${72.17}_{1.14}$ & ${76.75}_{0.87}$ & ${71.80}_{0.79}$ & ${76.79}_{0.93}$ & ${67.99}_{0.63}$ \\
    ERM(ES) & $\text{DeBERTaV3}_{\text{Large}}$ & - & 69.44 & ${79.01}_{4.44}$ & ${66.86}_{15.72}$ & ${64.37}_{14.58}$ & ${72.36}_{9.20}$ & ${68.22}_{7.24}$ & ${73.65}_{5.12}$ & ${66.44}_{6.05}$ \\
    DM & $\text{DeBERTaV3}_{\text{Large}}$ & Random & 100.00 & ${76.02}_{0.62}$ & ${64.53}_{4.45}$ & ${63.44}_{2.09}$ & ${73.13}_{2.12}$ & ${64.23}_{4.44}$ & ${71.84}_{3.41}$ & ${64.03}_{1.90}$ \\
    \midrule
    \multicolumn{11}{c}{Training Dynamics Transferability: Across Different Model Sizes} \\ 
    \midrule
    DM & $\text{DeBERTaV3}_{\text{Large}}$ & $\text{DeBERTaV3}_{\text{Large}}$ & 200.00 & ${80.52}_{0.89}$ & ${79.57}_{0.83}$ & ${74.34}_{2.30}$ & ${76.47}_{1.33}$ & ${71.19}_{1.38}$ & ${75.96}_{2.35}$ & ${68.19}_{0.90}$ \\
    DM & $\text{DeBERTaV3}_{\text{Large}}$ & $\text{DeBERTaV3}_{\text{Small}}$ & 114.47 & ${81.27}_{1.11}$ & ${76.46}_{2.65}$ & ${74.71}_{1.70}$ & ${77.00}_{0.82}$ & ${72.82}_{1.36}$ & ${76.99}_{0.97}$ & ${68.79}_{1.49}$ \\
    DM & $\text{DeBERTaV3}_{\text{Large}}$ & $\text{DeBERTaV3}_{\text{Base}}$ & 128.29 & ${80.63}_{0.57}$ & ${78.98}_{1.18}$ & ${74.46}_{1.94}$ & ${75.87}_{0.35}$ & ${73.37}_{2.89}$ & ${77.26}_{0.26}$ & ${67.57}_{0.83}$ \\
    \midrule
    \multicolumn{11}{c}{Training Dynamics Transferability: Across Different Pretraining Methods} \\ 
    \midrule
    DM & $\text{DeBERTaV3}_{\text{Large}}$ & $\text{ELECTRA}_{\text{Small}}$ & 104.61 & ${76.69}_{0.64}$ & ${73.12}_{2.61}$ & ${71.64}_{4.16}$ & ${76.91}_{1.55}$ & ${70.42}_{2.36}$ & ${76.68}_{0.99}$ & ${68.77}_{2.39}$ \\
    DM & $\text{DeBERTaV3}_{\text{Large}}$ & $\text{ELECTRA}_{\text{Base}}$ & 136.18 & ${80.38}_{1.28}$ & ${78.55}_{2.05}$ & ${76.75}_{2.22}$ & ${77.55}_{1.67}$ & ${73.86}_{2.45}$ & ${77.11}_{0.31}$ & ${69.35}_{1.65}$ \\
    DM & $\text{DeBERTaV3}_{\text{Large}}$ & $\text{RoBERTa}_{\text{Large}}$ & 216.78 & ${79.93}_{0.94}$ & ${77.33}_{1.71}$ & ${75.09}_{1.68}$ & ${76.99}_{1.53}$ & ${72.36}_{0.54}$ & ${76.96}_{1.19}$ & ${67.66}_{1.84}$ \\
    DM & $\text{DeBERTaV3}_{\text{Large}}$ & $\text{BERT}_{\text{Large}}$ & 213.49 & ${80.51}_{0.92}$ & ${80.28}_{3.19}$ & ${77.53}_{2.00}$ & ${78.72}_{1.55}$ & ${73.43}_{2.24}$ & ${76.94}_{1.39}$ & ${68.54}_{1.32}$ \\
    \midrule
    \multicolumn{11}{c}{FTFT: Efficient Reference Models + Aggressive Early Stopping} \\ 
    \midrule
    FTFT & $\text{DeBERTaV3}_{\text{Large}}$ & $\text{DeBERTaV3}_{\text{Small}}$ & 78.36 & ${80.20}_{0.89}$ & ${78.13}_{1.98}$ & ${74.85}_{2.16}$ & ${77.59}_{1.56}$ & ${72.44}_{0.45}$ & ${77.44}_{1.38}$ & ${69.28}_{1.38}$ \\
    FTFT & $\text{DeBERTaV3}_{\text{Large}}$ & $\text{DeBERTaV3}_{\text{Base}}$ & 106.07 & ${79.97}_{1.17}$ & ${77.27}_{3.93}$ & ${74.07}_{1.93}$ & ${76.12}_{0.86}$ & ${74.38}_{2.68}$ & ${76.91}_{1.85}$ & ${68.90}_{1.07}$ \\
    FTFT & $\text{DeBERTaV3}_{\text{Large}}$ & $\text{ELECTRA}_{\text{Base}}$ & 108.40 & ${79.40}_{0.92}$ & ${80.10}_{2.05}$ & ${76.26}_{1.82}$ & ${79.07}_{1.56}$ & ${75.69}_{1.46}$ & ${77.93}_{0.50}$ & ${69.44}_{1.65}$ \\
    \bottomrule
\end{tabular}
\caption{
    Our results of $\text{DeBERTaV3}$ as the main model on HSD (measured by F1 scores), 
    which consist of four parts:
    (1) Baselines: 
        $\text{DeBERTaV3}$ of different sizes trained using ERM, 
        $\text{DeBERTaV3}_{\text{Large}}$ trained using ERM with early stopping (ERM(ES)), and 
        $\text{DeBERTaV3}_{\text{Large}}$ trained using random DM (random 33\% of the training data); 
    (2) Training dynamics transferability across different sizes: 
        training $\text{DeBERTaV3}_{\text{Large}}$ as the main model, 
        using DMs constructed by different sizes of $\text{DeBERTaV3}$ as reference models;
    (3) Training dynamics transferability across different pretraining methods:
        training $\text{DeBERTaV3}_{\text{Large}}$ as the main model, 
        using DMs constructed by different pretraining methods as reference models, 
        including $\text{ELECTRA}_{\text{Small}}$, $\text{ELECTRA}_{\text{Base}}$,
        $\text{TinyBERT}$, and $\text{RoBERTa}$; 
    (4) FTFT: 
        training $\text{DeBERTaV3}_{\text{Large}}$ using our approach FTFT, 
        with DMs constructed by different reference models, as well as aggressive early stopping.
    Ori/Pert and R2--R4 in DynaHate refer to different rounds of 
    collected \textbf{Ori}ginal and \textbf{Pert}urbed data. 
    Compute refers to the relative training computational cost 
    compared to training $\text{DeBERTaV3}_{\text{Large}}$ using ERM.
    We observe that: 
    (1) Training dynamics are transferrable across different sizes and pretraining methods,
    as constructing DMs using different reference models results in comparable performance; 
    (2) FTFT achieves consistent robustness improvements over ERM, 
    while maintaining or lowering the training cost. 
    (3) FTFT enhances efficiency more when the optimal length of training is longer 
    (ERM only trains 1.6k steps on CAD). 
}
\label{tab:hsd_overall}
\end{sidewaystable*}

\clearpage
\begin{sidewaystable*}
\centering

\begin{tabular}{lll|c|cccccccc}
    \toprule
    Method & Main Model & Ref. Model & Compute & CAD (Val.) & \multicolumn{6}{c}{DynaHate (Test)} \\
    &  &  &  & - & Ori-R2 & Ori-R3 & Ori-R4 & Pert-R2 & Pert-R3 & Pert-R4 \\
    \midrule
    \multicolumn{11}{c}{ERM Performance: Capabilities of Various Models} \\ 
    \midrule
    \rowcolor[HTML]{C0C0C0} 
    ERM & $\text{TinyBERT}$ & - & 1.45 & ${63.85}_{0.95}$ & ${36.32}_{0.91}$ & ${41.68}_{2.40}$ & ${44.93}_{0.68}$ & ${31.99}_{1.62}$ & ${44.28}_{2.49}$ & ${43.12}_{1.17}$ \\
    \rowcolor[HTML]{C0C0C0} 
    ERM & $\text{ELECTRA}_{\text{Small}}$ & - & 4.61 & ${71.55}_{0.86}$ & ${47.36}_{2.31}$ & ${47.83}_{0.90}$ & ${55.23}_{2.48}$ & ${48.30}_{2.28}$ & ${57.18}_{2.28}$ & ${55.37}_{1.95}$ \\
    ERM & $\text{ELECTRA}_{\text{Base}}$ & - & 36.18 & ${76.62}_{1.23}$ & ${59.43}_{2.73}$ & ${57.86}_{2.35}$ & ${65.75}_{1.59}$ & ${57.49}_{3.21}$ & ${65.15}_{0.56}$ & ${63.48}_{1.33}$ \\
    ERM & $\text{DeBERTaV3}_{\text{Small}}$ & - & 14.47 & ${75.92}_{1.20}$ & ${50.63}_{2.40}$ & ${52.31}_{2.35}$ & ${61.37}_{1.30}$ & ${57.98}_{2.19}$ & ${65.42}_{0.83}$ & ${61.42}_{0.45}$ \\
    ERM & $\text{DeBERTaV3}_{\text{Base}}$ & - & 28.29 & ${77.70}_{1.01}$ & ${55.97}_{0.68}$ & ${58.88}_{2.57}$ & ${63.79}_{1.67}$ & ${59.76}_{0.85}$ & ${66.63}_{1.39}$ & ${61.07}_{0.86}$ \\
    ERM & $\text{BERT}_{\text{Large}}$ & - & 113.49 & ${77.57}_{0.77}$ & ${55.80}_{3.04}$ & ${61.62}_{3.00}$ & ${64.60}_{3.53}$ & ${61.82}_{0.78}$ & ${64.69}_{1.26}$ & ${65.00}_{2.48}$ \\
    ERM & $\text{RoBERTa}_{\text{Large}}$ & - & 116.78 & ${78.92}_{0.94}$ & ${64.68}_{3.82}$ & ${65.95}_{1.10}$ & ${68.65}_{1.55}$ & ${63.73}_{0.55}$ & ${70.39}_{0.46}$ & ${64.27}_{1.50}$ \\
    \midrule
    \multicolumn{11}{c}{When Do Transfers Fail: Using Reference Models of Different Capabilities} \\ 
    \midrule
    \rowcolor[HTML]{C0C0C0} 
    DM & $\text{DeBERTaV3}_{\text{Large}}$ & $\text{TinyBERT}$ & 101.45 & ${77.97}_{1.07}$ & ${68.51}_{2.47}$ & ${68.89}_{1.73}$ & ${73.99}_{1.23}$ & ${65.60}_{3.96}$ & ${73.22}_{2.16}$ & ${66.36}_{0.94}$ \\
    \rowcolor[HTML]{C0C0C0} 
    DM & $\text{DeBERTaV3}_{\text{Large}}$ & $\text{ELECTRA}_{\text{Small}}$ & 104.61 & ${76.69}_{0.58}$ & ${73.12}_{2.36}$ & ${71.63}_{3.76}$ & ${76.91}_{1.41}$ & ${70.42}_{2.13}$ & ${76.68}_{0.89}$ & ${68.77}_{2.16}$ \\
    DM & $\text{DeBERTaV3}_{\text{Large}}$ & $\text{ELECTRA}_{\text{Base}}$ & 136.18 & ${80.38}_{1.16}$ & ${78.55}_{1.86}$ & ${76.75}_{2.01}$ & ${77.55}_{1.51}$ & ${73.86}_{2.21}$ & ${77.11}_{0.28}$ & ${69.35}_{1.49}$ \\
    DM & $\text{DeBERTaV3}_{\text{Large}}$ & $\text{DeBERTaV3}_{\text{Small}}$ & 114.47 & ${81.27}_{1.00}$ & ${76.46}_{2.39}$ & ${74.71}_{1.54}$ & ${77.00}_{0.74}$ & ${72.82}_{1.23}$ & ${76.99}_{0.87}$ & ${68.79}_{1.35}$ \\
    DM & $\text{DeBERTaV3}_{\text{Large}}$ & $\text{DeBERTaV3}_{\text{Base}}$ & 128.29 & ${80.63}_{0.52}$ & ${78.98}_{1.07}$ & ${74.46}_{1.75}$ & ${75.87}_{0.31}$ & ${73.37}_{2.61}$ & ${77.25}_{0.24}$ & ${67.57}_{0.75}$ \\
    DM & $\text{DeBERTaV3}_{\text{Large}}$ & $\text{BERT}_{\text{Large}}$ & 213.49 & ${80.51}_{0.92}$ & ${80.28}_{3.19}$ & ${77.53}_{2.00}$ & ${78.72}_{1.55}$ & ${73.43}_{2.24}$ & ${76.94}_{1.39}$ & ${68.54}_{1.32}$ \\
    DM & $\text{DeBERTaV3}_{\text{Large}}$ & $\text{RoBERTa}_{\text{Large}}$ & 216.78 & ${79.93}_{0.85}$ & ${77.33}_{1.54}$ & ${75.09}_{1.52}$ & ${76.99}_{1.38}$ & ${72.36}_{0.49}$ & ${76.96}_{1.07}$ & ${67.66}_{1.67}$ \\
    \midrule
    \rowcolor[HTML]{C0C0C0} 
    DM & $\text{ELECTRA}_{\text{Large}}$ & $\text{TinyBERT}$ & 111.64 & ${76.64}_{0.58}$ & ${67.59}_{1.86}$ & ${65.36}_{1.97}$ & ${71.35}_{1.84}$ & ${64.85}_{2.66}$ & ${70.37}_{0.74}$ & ${67.77}_{1.57}$ \\
    \rowcolor[HTML]{fc7f03} 
    DM & $\text{ELECTRA}_{\text{Large}}$ & $\text{ELECTRA}_{\text{Small}}$ & 114.80 & ${71.40}_{11.28}$ & ${57.75}_{16.82}$ & ${57.93}_{14.85}$ & ${63.25}_{17.15}$ & ${57.05}_{17.35}$ & ${63.60}_{14.16}$ & ${62.05}_{12.46}$ \\
    DM & $\text{ELECTRA}_{\text{Large}}$ & $\text{ELECTRA}_{\text{Base}}$ & 146.38 & ${78.61}_{0.31}$ & ${76.33}_{1.04}$ & ${69.80}_{2.10}$ & ${74.23}_{1.65}$ & ${69.12}_{0.95}$ & ${71.72}_{1.39}$ & ${69.23}_{1.22}$ \\
    DM & $\text{ELECTRA}_{\text{Large}}$ & $\text{DeBERTaV3}_{\text{Small}}$ & 124.67 & ${78.37}_{0.81}$ & ${76.39}_{0.75}$ & ${68.51}_{0.77}$ & ${72.12}_{1.87}$ & ${68.81}_{0.80}$ & ${72.58}_{0.97}$ & ${68.83}_{0.99}$ \\
    \rowcolor[HTML]{fc7f03}
    DM & $\text{ELECTRA}_{\text{Large}}$ & $\text{DeBERTaV3}_{\text{Base}}$ & 138.49 & ${74.87}_{3.75}$ & ${59.49}_{20.98}$ & ${58.43}_{13.10}$ & ${65.65}_{9.34}$ & ${57.63}_{14.30}$ & ${65.20}_{9.59}$ & ${60.47}_{10.88}$ \\
    DM & $\text{ELECTRA}_{\text{Large}}$ & $\text{BERT}_{\text{Large}}$ & 223.68 & ${78.38}_{1.27}$ & ${76.69}_{3.87}$ & ${71.80}_{1.82}$ & ${74.63}_{1.59}$ & ${71.62}_{1.36}$ & ${73.23}_{1.37}$ & ${69.09}_{0.57}$ \\
    \rowcolor[HTML]{fc7f03}
    DM & $\text{ELECTRA}_{\text{Large}}$ & $\text{RoBERTa}_{\text{Large}}$ & 226.97 & ${76.64}_{3.91}$ & ${66.46}_{10.28}$ & ${65.42}_{1.06}$ & ${72.47}_{2.36}$ & ${66.79}_{4.70}$ & ${71.58}_{1.91}$ & ${66.45}_{1.27}$ \\
    \midrule
    \rowcolor[HTML]{C0C0C0} 
    DM & $\text{DeBERTaV3}_{\text{Base}}$ & $\text{TinyBERT}$ & 29.74 & ${75.00}_{0.01}$ & ${53.00}_{0.06}$ & ${58.00}_{0.06}$ & ${63.00}_{0.04}$ & ${57.00}_{0.05}$ & ${66.00}_{0.03}$ & ${60.00}_{0.04}$ \\
    DM & $\text{DeBERTaV3}_{\text{Base}}$ & $\text{DeBERTaV3}_{\text{Small}}$ & 42.76 & ${77.00}_{0.02}$ & ${64.00}_{0.03}$ & ${67.00}_{0.02}$ & ${70.00}_{0.02}$ & ${64.00}_{0.03}$ & ${69.00}_{0.03}$ & ${63.00}_{0.02}$ \\
    \bottomrule
\end{tabular}

\caption{
    We use reference models of different capabilities (measured by their ERM F1 scores) 
    to construct DMs, 
    and use them to train main models. 
    The gray-shaded rows are the (1) main models in unsuccessful transfers 
    and (2) their corresponding reference models. 
    The orange-shaded rows are the rows with very high standard deviations: 
    we observe that \textbf{fine-tuning $\text{ELECTRA}_{\text{Large}}$ on small datasets like CAD is very unstable, 
    and often produces failed runs}. 
    Following \citet{mosbach2021on} we removed the runs with ID performance worse than the majority classifier: 
    however, there are still some runs with slightly better ID performance than the majority classifier, 
    but with diverged loss or fluctuating loss after the first a few training steps 
    (i.e. the loss curve going up, or the loss curve being almost flat), 
    and are significantly worse than the other runs. 
    \colorbox[HTML]{fc7f03}{These ``almost failed runs'' cause the high standard deviations in the orange-shaded rows.}
    We therefore \textbf{exclude these runs in our analyses in \S\ref{subsec:how_efficient_can_we_be}}. 
    Successful transfer requires the reference model to be reasonably strong:
    reference models with clearly worse ID performance lead to 
    degraded OOD performance for the main models. 
}
\label{tab:how_efficient_can_we_be_hsd}
\end{sidewaystable*}

\end{document}